\def\year{2019}\relax
\documentclass[letterpaper]{article} 
\usepackage{aaai19}  
\usepackage{times}  
\usepackage{helvet} 
\usepackage{courier}  
\usepackage[hyphens]{url}  
\usepackage{graphicx} 
\usepackage{graphicx}  
\usepackage{bm}
\usepackage{amsfonts}
\usepackage{nccmath}
\usepackage{amsmath} 
\usepackage{amssymb}  
\usepackage{multirow}
\usepackage{dsfont} 
\usepackage{resizegather}

\usepackage{graphics} 
\usepackage{epsfig} 
\usepackage{mathptmx} 
\usepackage{times} 
\usepackage{amsmath} 
\usepackage{amssymb}  
\usepackage{bm}
\usepackage{multirow}
\usepackage{dsfont} 

\usepackage{xcolor}
\usepackage{resizegather}
\usepackage{comment}
\usepackage{placeins}
\newcommand{\cmmnt}[1]{}
\usepackage{soul}

\title{Four-Arm Manipulation via  Feet Interfaces} 
\author{Jacob Hernandez\textsuperscript{\rm 1}, Walid Amanhoud\textsuperscript{\rm 2}, Ana\"is Haget\textsuperscript{\rm 2}, \\ \Large \textbf { Hannes Bleuler\textsuperscript{\rm 1}, Aude Billard\textsuperscript{\rm 2} and Mohamed Bouri\textsuperscript{\rm 1}}\\ 
	\textsuperscript{\rm 1}REHAssist Group, EPFL. Station 9, STI IMT MED, Lausanne Switzerland\\ 
	\textsuperscript{\rm 2}LASA Laboratory, EPFL. Station 9, STI IMT MEB, Lausanne Switzerland\\
	jacob.hernandezsanchez@epfl.ch 
}

\begin{document}

\maketitle
\thispagestyle{empty}
\pagestyle{empty}

\begin{abstract}

We seek to augment human manipulation by enabling humans to control two robotic arms in addition to their natural arms using their feet. Thereby, the hands are free to perform tasks of high dexterity, while the feet-controlled arms perform tasks requiring lower dexterity, such as supporting a load. The robotic arms are tele-operated through two foot interfaces that transmit translation and rotation to the end effector of the manipulator. Haptic feedback is provided for the human to perceive contact and change in load and to adapt the feet pressure accordingly.

Existing foot interfaces have been used primarily for a single foot control and are limited in range of motion and number of degrees of freedom they can control. This paper presents foot-interfaces specifically made for bipedal control, with a workspace suitable for two feet operation and in five degrees of freedom each. This paper also presents a position-force teleoperation controller based on Impedance Control modulated through Dynamical Systems for trajectory generation. Finally, an initial validation of the platform is presented, whereby a user grasps an object with both feet and generates various disturbances while the object is supported by the feet.

\end{abstract}

\setlength{\fboxsep}{0pt}%
\setlength{\fboxrule}{0pt}%

\section{Introduction}\label{section:introduction}


There is evidence that feet could potentially be good candidates for controlling robotic arms. Starting by studies in feet-computer interaction where the feet have been found appropriate for accurate and non-accurate spatial tasks \cite{HOFFMANN1991} \cite{Pakkanen2004}, and recently \cite{Abdi2016} found that having a mental representation of one foot as a third hand in a virtual environment can improve performance in cognitively demanding scenarios.  

\begin{figure}[t!]
	\centering
	\framebox{\parbox{3.3in}{
	        \centering
			\includegraphics[width=\columnwidth]{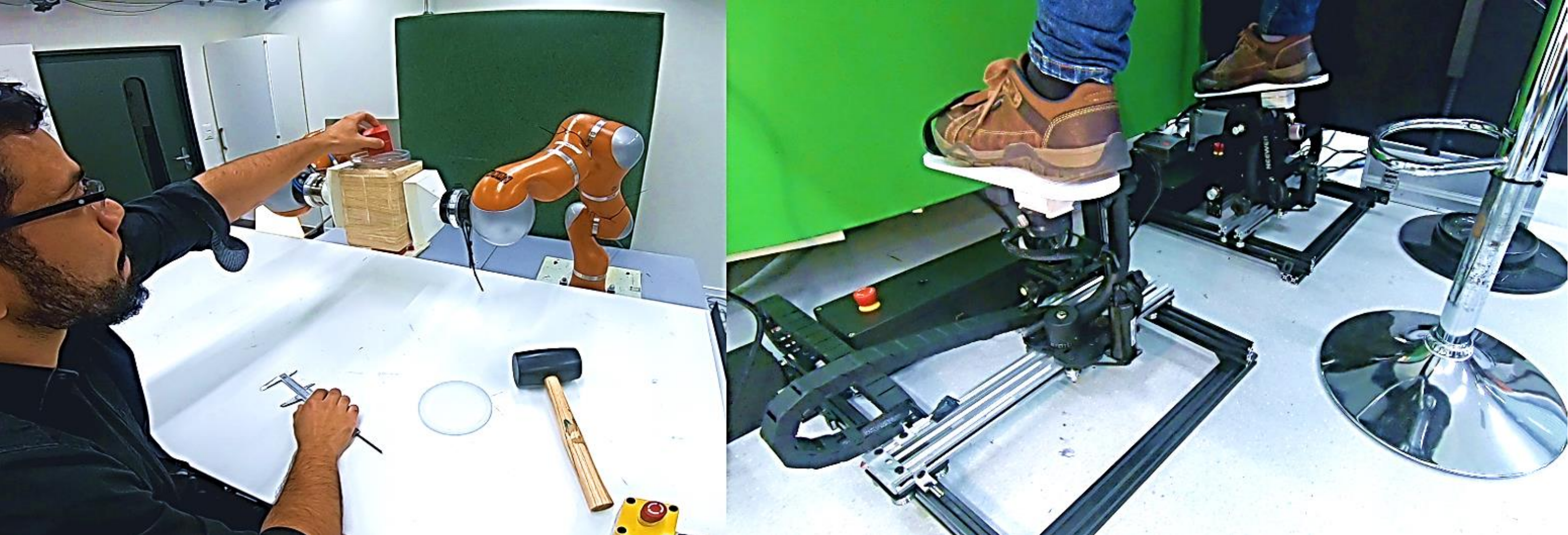}
			\caption{The user drives two robotic arms with the feet. Each robot is controlled by their ipsilateral foot in cartesian teleoperation. The forces measured from each robot are fed back to the user through haptic feedback.}
			\label{fig_Overall}
	}}
\end{figure}

We investigate the design of a feet-interface, namely an interface that can be operated by the two feet simultaneously and show how such an interface can be used to enable a four-handed telemanipulation (Fig. \ref{fig_Overall}).
 
Our use case contrasts other approaches for Supernumerary Robotic Limbs (SRL) as in {\cite{Llorens-Bonilla2012}}{\cite{Bonilla2014}}{\cite{Bright2017}}, in the fact that in our case the human has control over the artificial robotic arms using the natural dexterity of their feet. This control can potentially be within a spectrum from direct manipulation of the motor commands towards a shared autonomy to facilitate the task for the human. 

Not only does the interaction through feet leave the hands free to perform other tasks, but the haptic link (between human and robot) allows the human to supervise the desired motion and force of the task. This may be advantageous in human-robot collaborative scenarios where visual or verbal guidance may compromise the efficiency, responsiveness or the quality of the task (e.g. assisted surgery, complex assemblies, etc).


\section{Feet Interfaces} 

Unlike its hand counterpart, a foot interface has the additional challenge that the leg represents a considerable load, namely the weight of the leg, that depending on the body posture can vary from $18\%$ to $100\%$ of body mass in healthy adults (i.e. $\geqslant 15$ kg) \cite{Plagenhoef1983}. Since the high payload challenges the mechatronic design, it is not a surprise that even though there are multiple existing foot platforms reported in literature (for rehabilitation, locomotion, etc), not all of them are convenient for teleoperation using both feet. Many of them are cumbersome like \cite{Otis2008} \cite{Iwata2001} \cite{Yoon2006a} because of the big electromagnetic actuators to reflect high forces and to compensate the weight of the leg when moving up and down. Some machines are limited in degrees of freedom (from one to three) like \cite{Farjadian2014} \cite{Saglia2013} \cite{Wang2013} because they were designed for the ankle and not for motion of the leg. Moreover, many  of them are devoid of active haptic feedback like \cite{Paradiso2004} \cite{Rovers2005} \cite{Abdi2017a}. Finally, most of the surveyed interfaces employ parallel kinematics \cite{Girone2001} \cite{Saglia2013} \cite{Wang2013}, which despite being advantageous in terms of rigidity and low inertia, limits the workspace for linear motions with respect to the total footprints. 

Our contribution comes as a mechanical solution for feet to move in a large workspace relative to the footprint of the platforms. Also, since we are targeting the use of both feet simultaneously, feet should be as well able to move closely relative to each other. Additionally, we sought for a mechanical implementation that could facilitate ad hoc modifications of the workspace by choosing serial kinematics and using joints that could be easily constrained.

\begin{figure}[t!]
	\centering
	\framebox{\parbox{3.3in}{
	        \centering
			\includegraphics[width=\columnwidth]{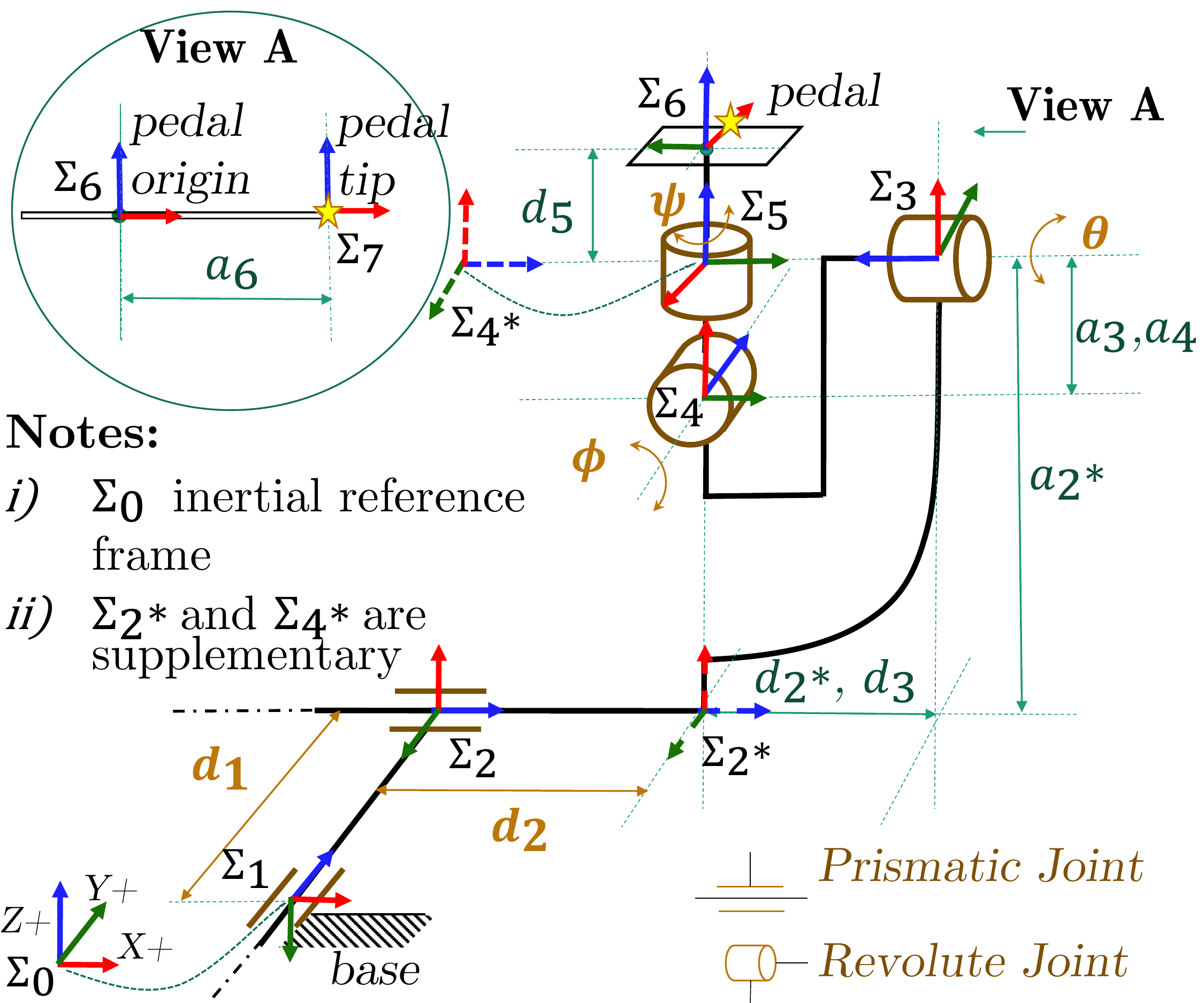}
			\caption{Denavit Hartenberg Kinematic Model of 5 DOF. Two prismatic joints drive linear motions and three rotational joints control the orientations. The motion command is in the frame of reference of the tip of the pedal ($\Sigma_7$). A small offset between the joints $3$ and $4$ (i.e. $a_3$) avoids the problem of Gimbal lock, therefore there is no singular configuration within the limits of the workspace.  }
			\label{fig_Model_FI}
	}}
\end{figure}

\subsection{Kinematic Model}

The kinematic model is illustrated in Fig \ref{fig_Model_FI}. We reduced the number of degrees of freedom to five to alleviate high torque requirements for compensating the inertial forces on the up-down motion of the legs.   

Let us consider the convention that the coordinates frames are defined as a set of orthonormal basis identified by $\Sigma_{\{.\}}$. 

The notation adopted in this paper for the kinematic formulation, escapes the conventional Denavit Hartenberg (DH) parameterization in that we are using intermediate supplementary frames (fixed joints) for convenience and clarity in the solution. This formulation specially allows to represent, in the kinematic chain, the three tait bryan angles using DH parameters.

$\Sigma_{0}$ is the inertial reference frame which is static or moving with constant velocity.

Let $^{i}T_{i+1} \in \mathds{R}^{4x4}$  be the coordinate transformation matrix between $\Sigma_{i+1}$ and $\Sigma_{i}$ consequence of the homogeneous transformations:
	
\begin{gather}
 \bm{^{i}T_{i+1}} = 
 {\begin{bmatrix}
	\bm{R_{z_i}}(\beta_i)  & \bm{p_{z_i}}(d_i) \\
	\bm{0^{1x3}} & 1
 \end{bmatrix}}
 {\begin{bmatrix}
 \bm{R_{x_{i+1}}}(\alpha_i)  & \bm{p_{x_{i+1}}}(a_i) \\
 \bm{0^{1x3}} & 1
 \end{bmatrix}}
\end{gather}
	
\noindent where $\bm{p_{w_{m}}}(.) \in \mathds{R}^ 3$ is a pure translation over an arbitraty axis $w$ of arbitraty frame $m$ and $\bm{R_{w_m}}(.) \in \mathds{R}^{3x3}$ is a pure rotation matrix around an arbitraty axis $w$ of arbitraty frame $m$. Following the convention,  $\beta_i$ and $d_i$ are the  angle and displacement in the $z_i$ axis, whereas $\alpha_i$ and $a_i$ are the angle and translation in the $x_{i+1}$ axis respectively.

Table \ref{tab_geom} presents the geometric parameters of the kinematic chain.
\begin{table}[htbp!]
	\caption{DH Parameters of the Kinematic Model}
	\begin{center}
		\begin{tabular}{|c|c|c|c|c|}
			\hline
			$\mathbf{^{i}T_{i+1}}$ & $\mathbf{\beta_i}$ & $\mathbf{d_{i}}$ & $\mathbf{\alpha_x}$ & $\mathbf{a_{i}}$  \\
			\hline
			${^{0}T_{1}} $ & $0$ & $0$  & $-\pi/2$ & $0$  \\
			$^{1}T_{2} $ &  $-\pi/2$ & $d_1$ & $-\pi/2$ & $0$  \\
			$^{2}T_{2^{*}} $ & $0$ & $d_2$ & $0$ & $0$ \\
			$^{2^{*}}T_{3} $ & $0$ & $d_{2^*}$& $\pi$  & $a_{2^{*}}$   \\
			$^{3}T_{4} $ & ${\theta}$ & $d_3$ & $-\pi/2$ & $-a_3$   \\
			$^{4}T_{4^{*}} $ & ${\phi}$ & $0$ & $-\pi/2$ & $a_4$   \\
			$^{4^{*}}T_{5} $ & $\pi/2$ & $0$  & $\pi/2$ & $0$  \\
			$^{5}T_{6} $ & $\pi + {\psi}$ & $d_5$ & $0$ & $0$   \\
			$^{6}T_{7} $ & $0$ & $0$ & $0$ & $a_6$   \\
			\hline
		\end{tabular}
		\begin{tabular}{c c c c c c c}
	    \\
		\end{tabular}
		\label{tab_geom}
	\end{center}
\end{table}
The forward kinematic model is obtained as follows:
\begin{equation}
\bm{^{0}T_{n} = \prod_{i=0}^{n-1} \left( ^{i}T_{i+1} \right)}
\end{equation}\label{eq_mgd}

\noindent Where $n$ is the number of coordinate frames, in this case 7, and $\prod(.)$ represents the pre-multiplication of successive transformation matrices. After computing the forward kinematics, the desired motion of the foot is taken from the cartesian coordinates of the frame at the tip of the pedal, namely $\bm{x^P} \in \mathds{R}^3 = \Sigma_{7}$ , can be described in the inertial reference frame by the following vector: 
\normalsize
\begin{gather}\label{eq_4}
	\begin {bmatrix}
		^{0}x_7 \\
		^{0}y_7 \\ 
		^{0}z_7  
	\end{bmatrix}
	=
	\begin{bmatrix}
	d_2 +(d_5 + a_3)  s_{\phi} - a_6  s_{\psi} c_{\phi}  \\
	d_1 + d_5  c_{\phi}s_{\theta} +a_6  {\left( c_{\theta} c_{\psi} + s_{\psi}s_{\theta}s_{\phi} \right)} + a_3 s_{\theta} {\left( c_{\phi} - 1 \right)} \\
	a_{2^{*}} - a_6 {\left( c_{\psi}s_{\theta} - c_{\theta}s_{\phi}s_{\psi} \right)} - a_3 c_{\theta} + d_5 c_{\theta}c_{\phi} + a_3 c_{\theta} c_{\phi}
	\end{bmatrix}	
\end{gather}
\normalsize
\noindent where
$ c_{\{{.}\}}$ and $ s_{\{.\}} $ correspond to the cosinus and sinus of the  angle ${.}$ respectively. 

The vector from (\ref{eq_4}) represents the task cartesian coordinates, and the geometric parameters $d_1$, $d_2$, $\theta$, $\psi$ and $\phi$ correspond to the generalized coordinates of the joint space.

\subsection{Resulting Workspace}

\begin{figure}[t!]
	\centering
	\framebox{\parbox{3.3in}{
	        \centering
		    \includegraphics[width=.9\columnwidth]{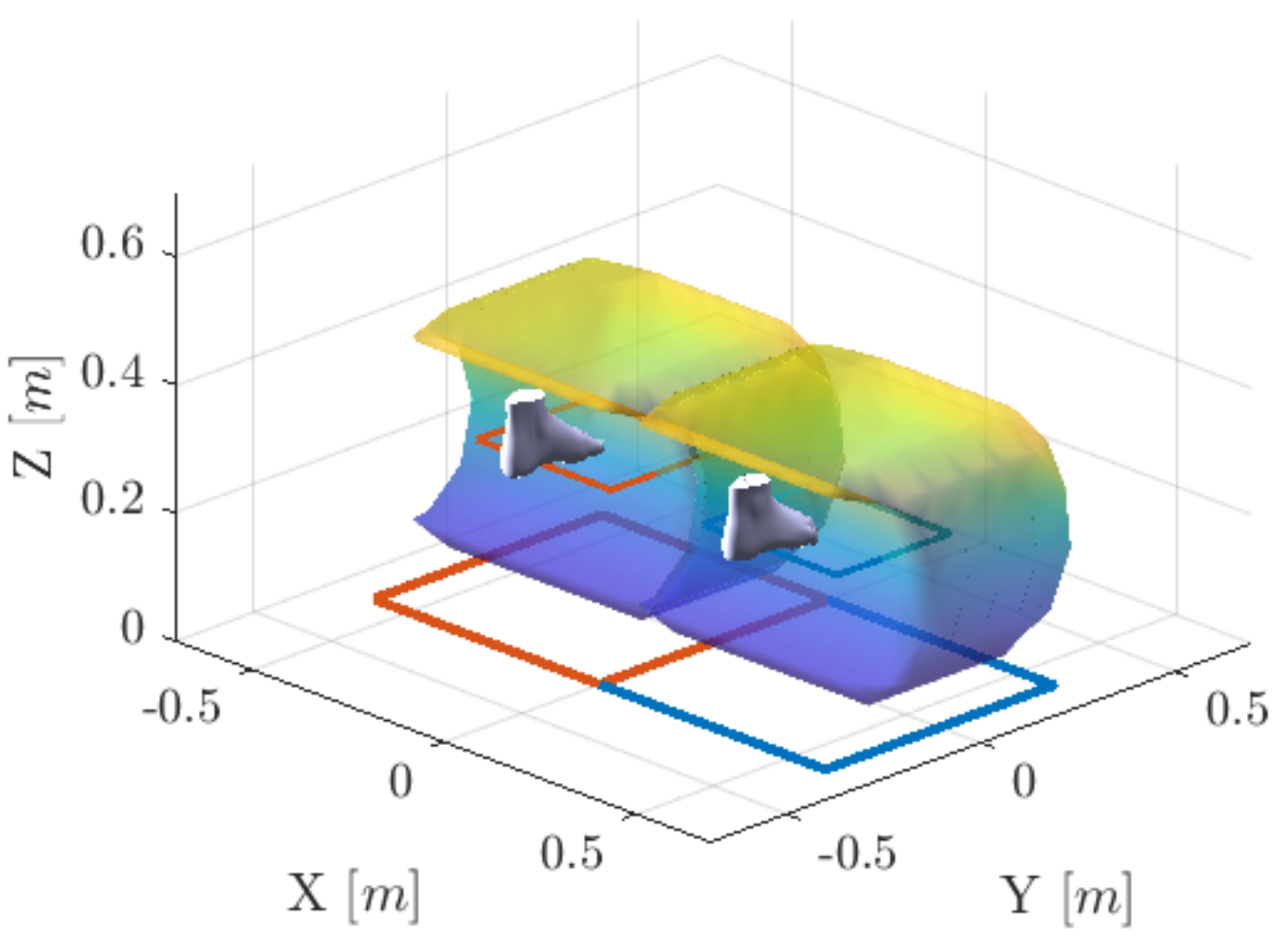}
			\caption{Illustration of the workspace of the feet pedals if the 5 DoF platforms were next to each other. The volume ($0.102 \text{ m}^3$ per foot) is computed from the forward kinematics on the pedal tips ($\Sigma_7$). The small rectangles ($0.350 \text{ m} \times 0.293 \text{ m}$) represent the linear range of motion in XY. To be compared with the net footprints of the platforms illustrated as big rectangles ($0.59\text{ m } \times 0.59\text{ m}$) in the base.}
			\label{fig_workspace}
	}}
\end{figure}

The workspace of the platform was computed from the forward kinematics of the Denavit-Hartenberg formulation, illustrated in Fig \ref{fig_workspace}, after considering the ranges of motion of the degrees of freedom ($d_1 \in [-175\text{ mm, } 175\text{ mm}]$, $d_2 \in [-146.5\text{ mm, }146.5\text{ mm}]$, $\theta \in [-80^\circ,80^\circ]$, $\phi \in [-25^\circ,  45^\circ]$ and $\psi \in [-45^\circ, 45^\circ]$ ) and the following geometrical parameters: $d_{2^{*}},d_3 = 170$ $\text{mm}$, $a_2 = 243$ $\text{mm}$,  $a_3,a_4 = 46$ $\text{mm}$, $d_5 = 40$ $\text{mm}$ and $a_6 = 300$ $\text{mm}$. These values were verified to go accordingly with the lower limbs effective workspace of an adult male of average height based on \cite{Pheasant1996}. The height ($a_2 + d_5$) was defined under technical constraints of the available hardware. 

\section{Hardware Implementation for Cartesian Control}

\begin{figure}[b!]
	\centering
	\framebox{\parbox{3.3in}{
	        \centering
			\includegraphics[width=\columnwidth]{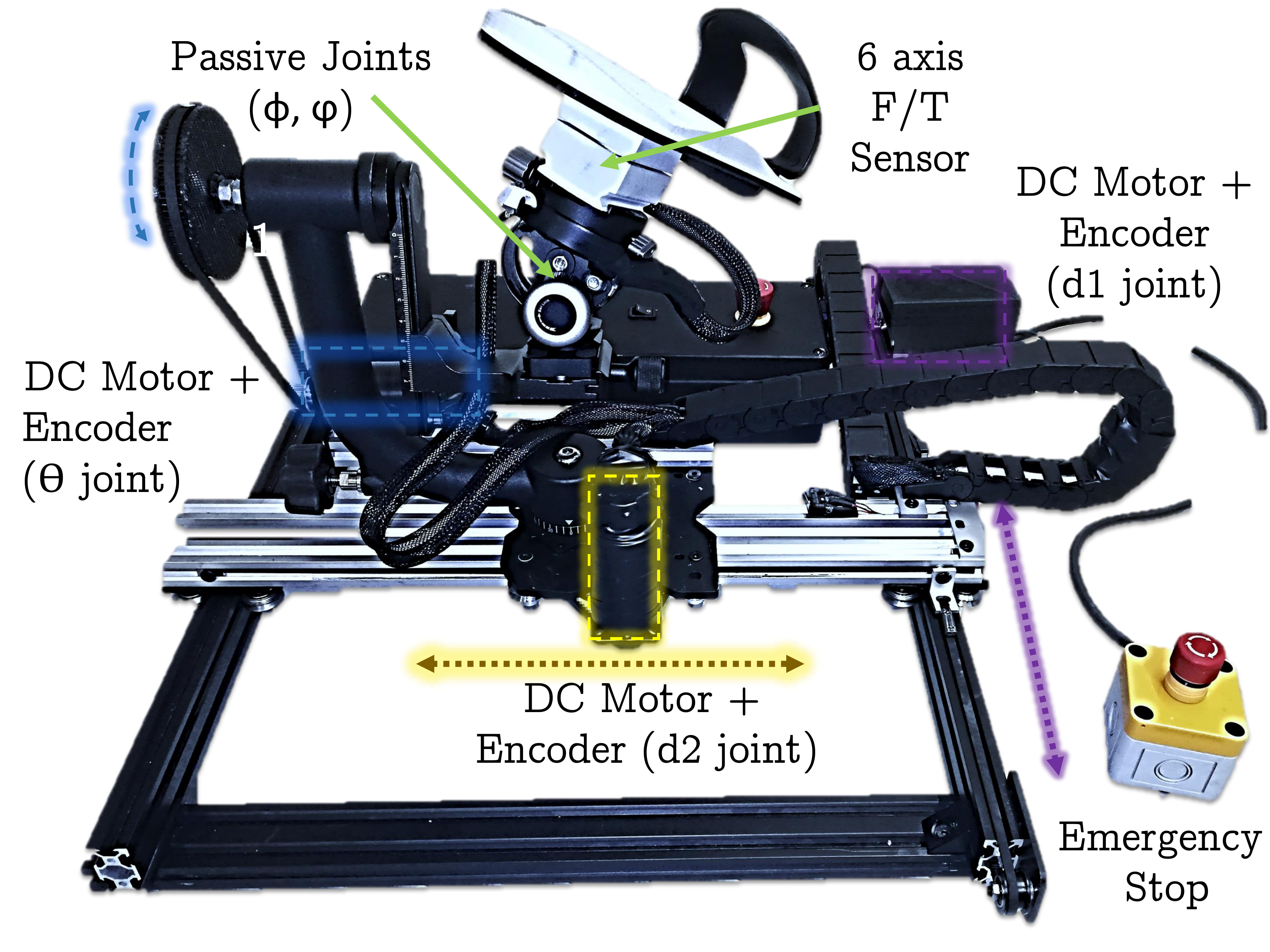}
			\caption{Illustration of the foot platform for linear cartesian control. It has 3D force feedback (X ($d_2$), Y ($d_1$),  $\theta$), highlighted with three different colors. DC motors provide force and the motion is measured with optical encoders. Two passive joints ($\phi\text{, }\gamma$) are fixed to a desired position measured with soft potentiometers. Each encoder is initialized using limit switches. All the axes are belt driven. The linear motions are achieved with v-slot aluminium profiles and adjustable rollers. The pulley for the pitch motion was custom made through 3D printing. A 6 axis ATI-Mini 40 Force/Torque Sensor is used to monitor the foot interaction forces.}
			\label{fig_FI_IO}
	}}
\end{figure}

\begin{figure}[t!]
	\centering
	\framebox{\parbox{3.3in}{
	        \centering
			\includegraphics[width=\columnwidth]{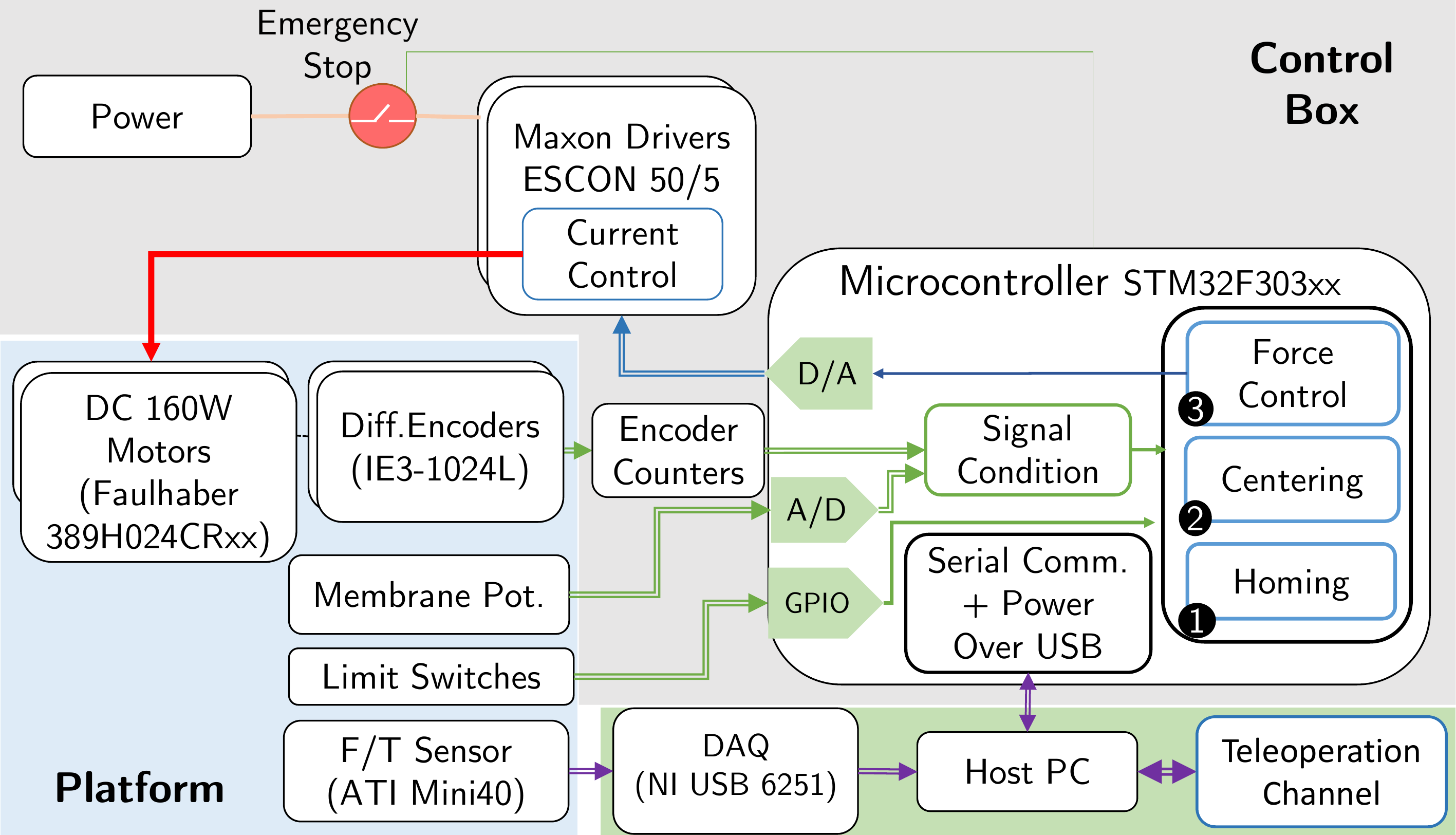}
			\caption{Block Diagram of Hardware \& Control Architecture. In the firmware implementation, homing and centering algorithms are followed by the implicit force control for teleoperation ( at $10$kHz). Moreover, the current control loop (PI) at $53.6$kHz. On the other hand, the teleoperation channel is done using Robot Operative System (ROS)} 
			\label{fig_hardware}
	}}
\end{figure}

The first demonstration simplifies the task to a linear cartesian teleoperation (3D). We decided to build a platform, see Fig \ref{fig_FI_IO}, following the kinematics presented in Fig \ref{fig_Model_FI}, but blocking the last two passive joints. Hence, each foot could control and receive feedback in 3 DoF. The future implementation will include 5D active force feedback.

The final specifications of the platform are listed in table \ref{tab_spec}. The ranges of motion were informed on available bio-mechanical data of the ankle (c.f \cite{Siegler1988}\cite{Dettwyler2004}). For the possible rotations, we allow a higher ROM than the anatomical constrains of the ankle  given the added extra mobility  acquired when engaging the movement of the entire leg.

Similarly, the dimensioning of the motors was based on the known psychophysics of perceivable forces of the foot's plantar/dorsiflexion. \cite{Southall1985} studied perception of resistive forces in a vehicle pedal (felt at the tip of the pedal around a lever arm). Results indicate that in ranges of background forces from $89$ to $445$ N, a Weber fraction  of $7\%$ should be applied for a difference to be detected by $50\%$ of the population. Similarly, \cite{Abbink} was found that footwear and frequency of duration affect the perception of active force variations, in a vehicle pedal, with background force of $25$ N . Results indicate that the just noticeable difference (JND) of a signal at $1$ Hz, when wearing socks, is $7$ N. This agrees with \cite{Ichinose2013} that report a JND of $7.4$ N in a study of driving assistance through pedal reaction force control. 

A recent study by \cite{Geitner2018} for determination of influence of footwear, pulse duration and amplitude, suggested that a good force reflection should span from $9$ to $18$ N to be comfortably detected. This agrees with  \cite{Abbink} and also \cite{EDWORTHY1995} that described that higher intensities than $18$N startle the driver. 

Consequently, we based our design on the recommended values for force reflection in plantar/dorsiflexion, expecting similar perception capabilities in the other foot rotations. 

Note that we envisioned the human to be in a sitting position as opposed to standing. We assumed that when sitting there is a greater body balance to move both hands and feet in multiple degrees of freedom.

Regarding the design of the mechanical structure, it consists in aluminium frames with v-groove (V-Slot $40x20x500 \text{ mm}^3$ ) that enable  self-centering smooth linear motions by the use of wheeled supported gantry plates moved by timing belt-pulley transmission (pulley radius of $9.15\text{ mm}$). For the rotary motion of the joint $\theta$ (See Fig. \ref{fig_Model_FI}), a bigger pulley was manufactured to amplify the torque $4.47$ times.

As illustrated in Fig. \ref{fig_FI_IO} the mechatronic design is comprised of a series of sensors and actuators for the motion input and the force reflection. DC $160 W$ motors (Faulhaber 38H024CRxx) are driven by servo-controller (MAXON-ESCON 50/5) and measured with incremental differential encoders (IE3-1024L). Moreover,  LS7366-based $32 bit$ encoder counters are communicated with the micro controller via serial peripheral interface (SPI).  The quadrature encoders are responsible for angle measurement of the actuated joints, whereas the passive joints are measured by using membrane potentiometers (Spectra Symbol SP-L-0100-103-3\%-RH). Then limit switches are implemented for reset of the values measured by the incremental encoders (homing). On the other hand, the membrane potentiometers provide absolute angle estimation. A six-axis force/torque (F/T) sensor (ATI Mini 40) is used to measure the interaction forces between the platform and the foot. 
The control is performed using an ARM Cortex M4-based micro-controller (STM32f303xx).

Relevant information about the control and hardware architectures can be appreciated in Fig.\ref{fig_hardware}.

\begin{table}[t!]
	\caption{Specifications of Foot Platform}
	\begin{center}
	\resizebox{\columnwidth}{!}{%
		\begin{tabular}{|c|c|c|}
			\hline
			\textbf{Metric} & \textbf{Design Specification} & \textbf{Value} \\
			\hline
			\multirow{2}{*}{Size} & Height& $0.365$ m\\
			& Footprint & $0.590$ m $\mathrm{x}$ $0.590$ m\\
			\hline
			\multirow{2}{*}{Range of Motion} & X $(d2)$ $\times$ Y $(d1)$  & $0.350 \times 0.293$  m$^2$\\
			& Pitch $(\theta)$ & $\pm80^{\circ}$\\
			\hline
			Transmission & Reduction ratio & $4.47:1$ \\ 
			\hline
			\multirow{4}{*}{Nominal Wrench} & \multirow{2}{*}{Force X and Y} & $15.1$ N (Nominal) \\ & & $45.3$ N (Peak)\\
			& \multirow{2}{*}{Torque Pitch} & $0.641$ Nm (Nominal)\\& & $1.923$ Nm (Peak)\\
			\hline
			\multirow{1}{*}{Motion Sensing} & Linear ($d_1$\&$d_2$) & $56$ um \\
			Resolution & Angular ($\theta$) & $0.08^{\circ}$ \\
			\hline
			\multirow{1}{*}{F/T Sensing} & Force ($d_1$\& $d_2$) & $1/50$ N\\
			Resolution\cite{ATI_S2018} & Torques ($\theta$) & $1/2000$ Nm\\
			\hline 
		\end{tabular}
		\label{tab_spec}}
		\end{center}
\end{table}


\section{Robot Control}

\begin{figure}[b!]
	\centering
	\framebox{\parbox{3.3in}{
	        \centering
			\includegraphics[width=\columnwidth]{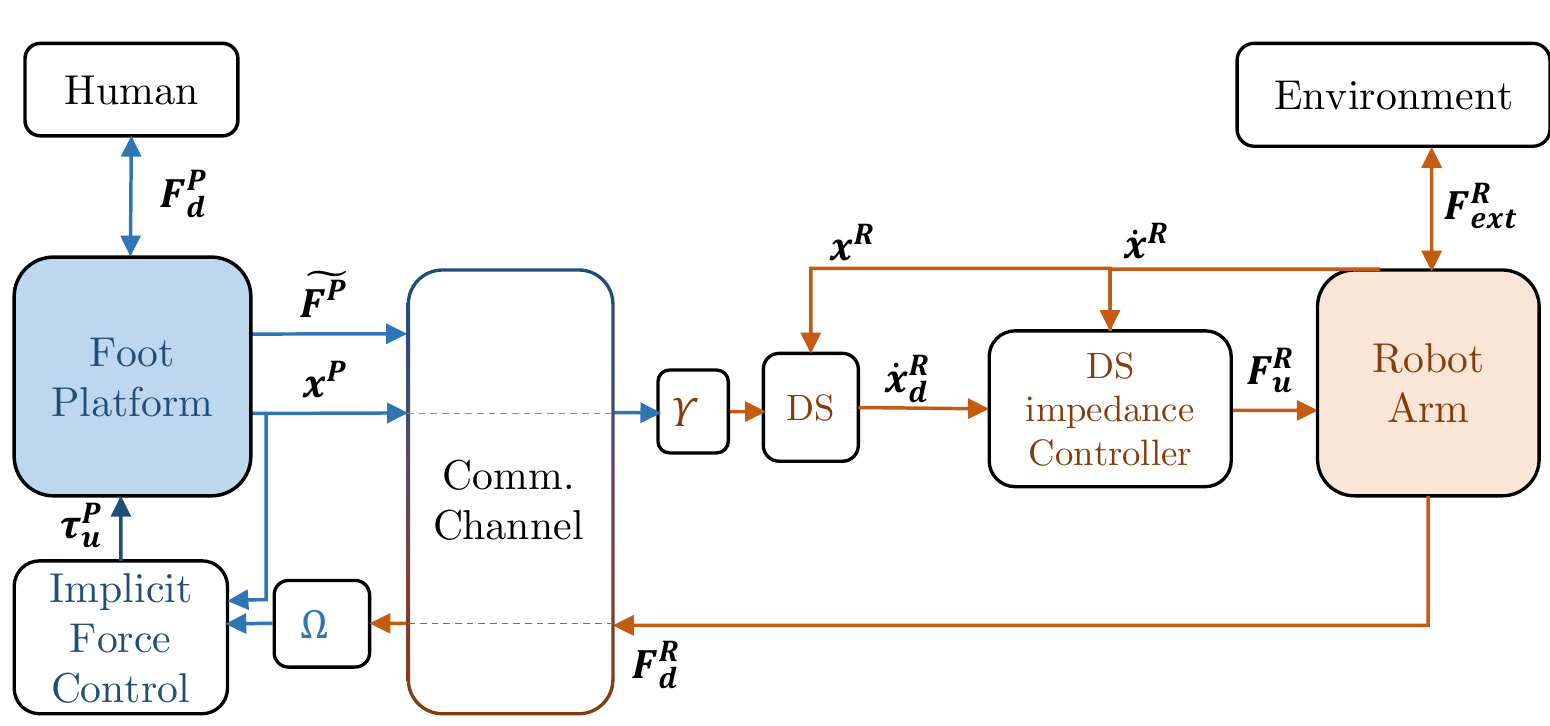}
			\caption{Position-Force DS-impedance based Teleoperation Architecture. The two colors represent the master and the tele-operated device.  The new variable introduced, $\bm{\tilde{F}^P}$ corresponds to the foot force that is monitored to check transparency and not used for closed loop control}
			\label{fig_teleoperation}
	}}
\end{figure}

The proposed control architecture of the position-force teleoperation is illustrated in Fig. \ref{fig_teleoperation}. We assume a constant negligible time delay in the communication channel, and also that the robotic arms are torque controlled.  

For clarity, we represent the variables related to the robot arm with the superscript $R$, while the variables of the foot platform with the superscript $P$.

For the \textit{telemanipulator side}, we start with the classical expression for the dynamics of a $n$ DOF manipulator in the three-dimensional cartesian space:
\small
\begin{gather} \label{eq_5}
\bm{B(x^R)\ddot{x}^R+C(x^R,\dot{x}^R)\dot{x}^R + G(x^R) = F_u^R + F_{ext}^R}	
\end{gather}
\normalsize
\noindent where $\bm{x^R} \in \mathds{R}^{3}$ 
denotes the position of the end effector, $\bm{B(x^R)} \in \mathds{R}^{3x3}$ the inertia matrix, $\bm{C(x^R,\dot{x}^R)\dot{x}^R} \in \mathds{R}^3$ the Centrifugal and Coriolis forces respectively, while $\bm{F_u^R} \in \mathds{R}^3$ and $\bm{F_{ext}^R}\in \mathds{R}^3$  correspond to the control and external forces respectively. The control force is obtained from an impedance controller taking the output of a time-invariant dynamical systems as reference velocity $\bm{\dot{x}^R_d}$, see \cite{Kronander2016}: 
\begin{equation}\label{eq_6}
\bm{F_u^R = D(x^R)(\dot{x}^R_d-\dot{x}^R)+\bm{G(x^R)}}
\end{equation} 
\noindent  where $\bm{G(x^R)}$ denotes the gravity compensation forces, $\bm{D(x^R)} \in \mathds{R}^{3\times3}$ is a varying damping matrix with positive eigenvalues $\lambda_{d1}$, $\lambda_{d1}$ and $\lambda_{d3}$, designed such that the first eigenvector is aligned with $\bm{\dot{x}^R_d}$. By manipulating the last two eigenvalues, one can selectively damp perturbations that are orthogonal to $\bm{\dot{x}^R_d}$, see \cite{Kronander2016}. This is advantageous to provide selective rigidity in directions that matter for the task (e.g. to the normal to the contact with the object) and hence to handle external disturbances in the teleoperation.  One clear advantage of this implementation is a safe physical robot interaction with unknown forces of the environment acting in directions not relevant to the task. A case in point is when we use both hands alongside the foot controlled telemanipulators to perform a supernumerary manipulation task; in such case, we can make the robot compliant to these exogenous forces when they are not aligned with the foot commands, and in consequence the feedback don't startle the user much.

To generate the desired robot velocity we use a linear dynamical system (DS) whose attractor is obtained by mapping the user foot's position to the robot's workspace: $ \bm{\dot{x}^R_d =  \Upsilon x^P-x^R}$, where $\bm{x^P} \in \mathds{R}^3$ is the foot position in the platform frame (in pedal tip $\Sigma_7$) and $\bm{\Upsilon} \in \mathds{R}^{3x3}$ is the  telefunctioning matrix mapping the platform's workspace to the robot one which takes into account rotations between both reference frames.

Finally, an orientation error measured in the end effector of the robotic arm (superscript E) is computed as as $\bm{\hat{R}= R_d R^{\intercal}}$ where $\bm{R=\left[x^E y^E z^E\right]} \in \mathds{R}^{3 \times 3}$ and  $\bm{R_d=\left[x_d^E y_d^E z_d^E\right]} \in \mathds{R}^{3 \times 3}$  are the axis-angle representations of the measured and the desired orientations for the end effector of the robot. The rotation target is met using a PD controller.

On the \textit{side of the foot master} device, we work on the joint space and establish the dynamics of the haptic device:
\scriptsize
\begin{gather} \label{eq_11}
\bm{b(q^P)\ddot{q}^P + c(q^P,\dot{q}^P) + \nu(q^P,\dot{q}^P) + g(q^P) } = \bm{\tau_{d}^P - \tau_{u}^P - \tau_n^P} 
\end{gather}
\normalsize
\noindent where $\bm{q^P} \in \mathds{R}^5$ represents the joint generalized coordinates,  $\bm{b(q^P)} \in \mathds{R}^{5x5}$ is the configuration-dependent inertia matrix, $\bm{\nu(q^P,\dot{q}^P)} \in \mathds{R}^5$  the non-linear velocity-dependent forces (frictions), $\bm{c(q^P)} \in \mathds{R}^5$ the centrifugal and coriolis and $\bm{g(q^P)} \in \mathds{R}^5$  the configuration dependent static forces (gravity). 
 $\bm{\tau_{u}^P}$ is the actuator commanded force expressed in the joint space,  $\bm{\tau_{n}^P}$ is the actuator disturbance projected in the joint space, and  $\bm{\tau_{d}^P}$ is the desired human input (interaction) force expressed in the joint space as $\bm{\tau_d^P=J_p^T F_d^P}$, where $\bm{F_d^P} \in \mathds{R}^3$ is the 3D cartesian force in the platform and $\bm{J_p} \in \mathds{R}^{3x5}$ is the translation submatrix of the geometrical Jacobian. Thus, $\bm{\dot{x}^P}=\bm{J_p}\bm{\dot{q}^P} $.

\begin{figure*}[t!]
	\centering
	\framebox{\parbox{.95\textwidth}{
			\centering
			\includegraphics[width=.95\textwidth
			]{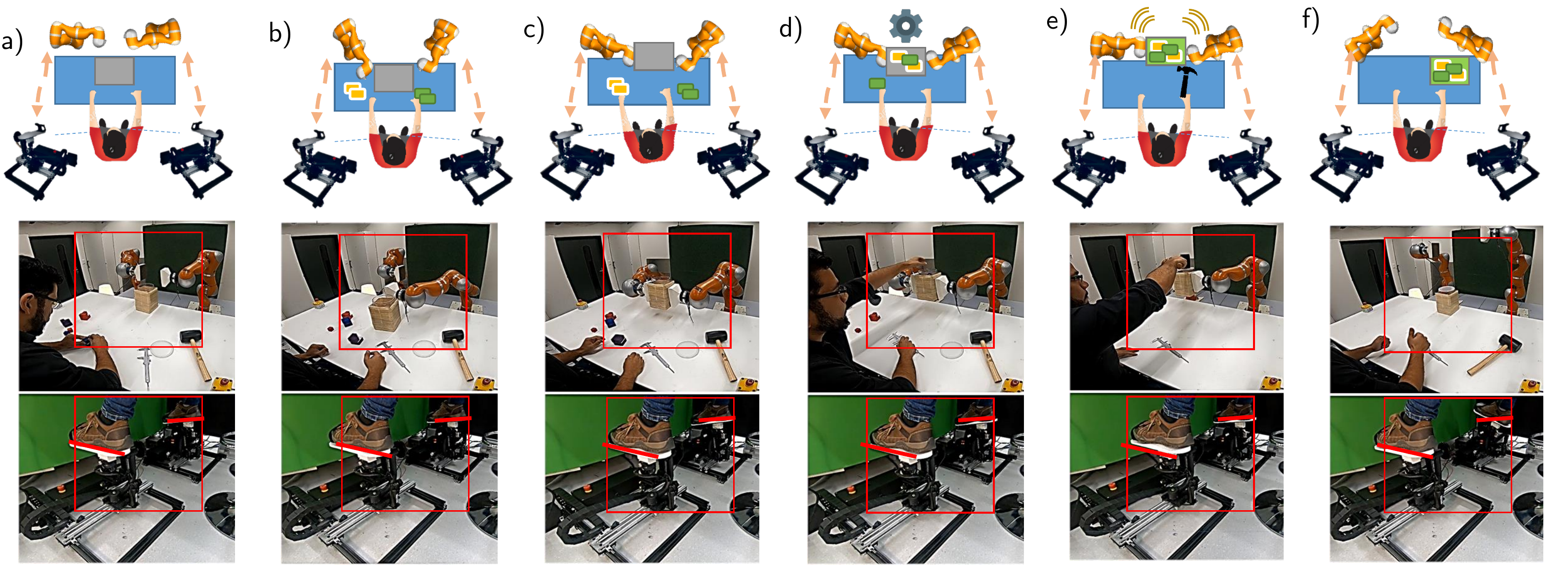}
			\caption{Illustrations of the bipedal telemanipulation to assist the work of the hands. The red lines are guides to understand the relative motion. Nevertheless, the video sequences provided as supplementary material are more telling that this description. Phases: a) No action is performed with the feet, while bimanual tasks are being done, b)  One robotic arm is used to retrieve a pieces container to assist the user in his task, c) To facilitate the task, the user grasps and lifts the container using the two foot-controlled robotic arms. d) The user adds the completed bundles in the container (working on container). e) The user performs a supernumerary task where, after placing a lid on the container, he is holding it with his feet while hammering. f) The user places the robotic arms far away from the working area since the tasks are completed (retreating).}
			\label{fig_snapshot}
	}}
    \framebox{\parbox{.95\textwidth}{
	        \centering
            \includegraphics[scale=0.5]{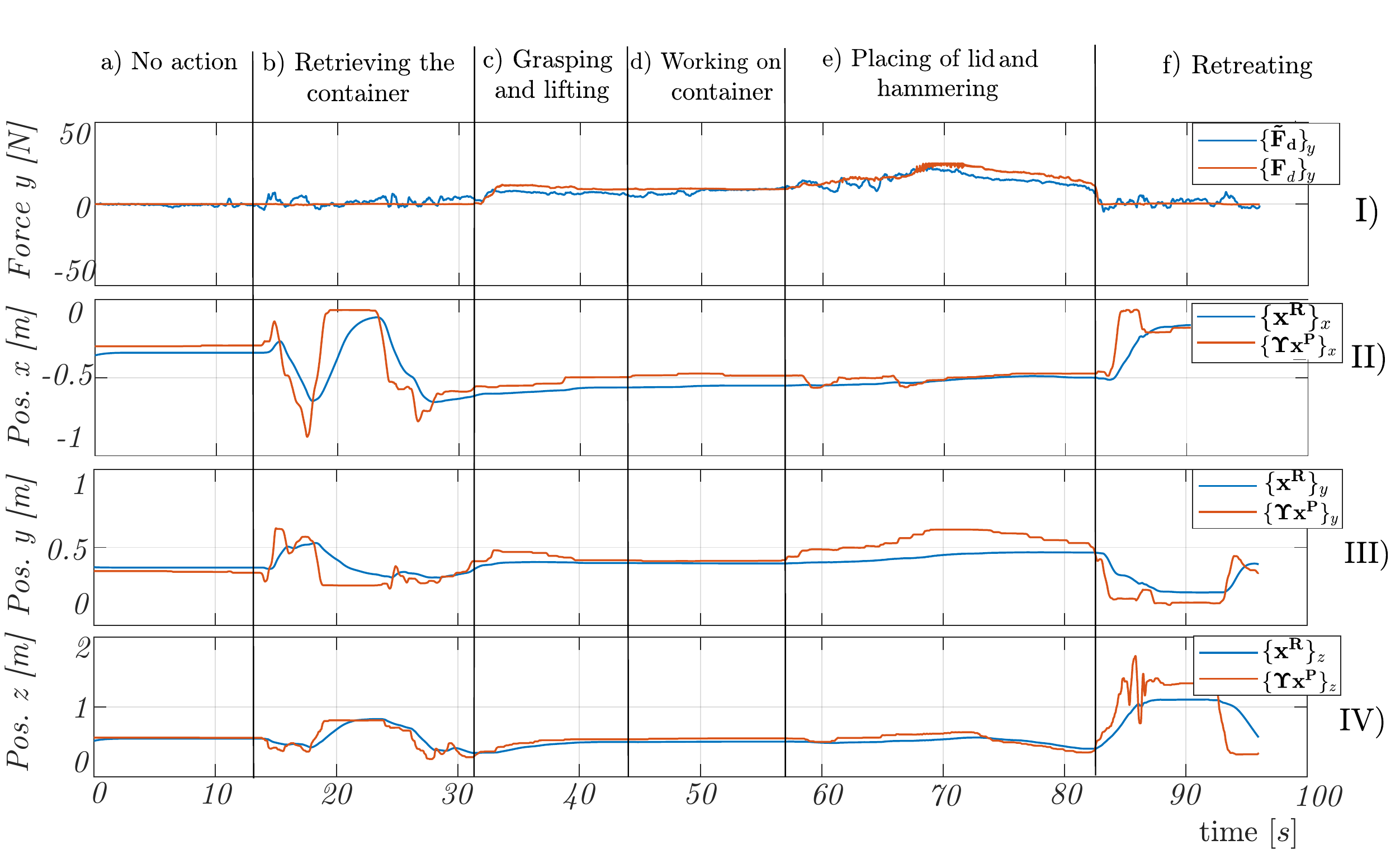}
    		\caption{Excerpt of Results for the Right Foot/Right Arm in the frame of the Right Arm $I.)$ Interaction force in the direction of grasping ($X$ in the Frame of the Platform, $Y$ in the Frame of the Kuka LWR), $\{\bm{\tilde{F}_d}\}_y$ (measured), and  $\{\bm{F}_d\}_y$ (desired) . $II.)$,  $III.)$ and $IV.)$ Plots in $X$, $Y$ and $Z$ of the robotic arm position from $\bm{\Upsilon x^P}$ (desired human input) and $\bm{x^R}$ (real measured position)}
    		\label{fig_human_response_task}
	}}
\end{figure*}

Furthermore, $\bm{F_d^P}$ is reflected from the telemanipulator as: $\bm{ F_d^P= \varOmega F_d^R}$, where $\bm{F_d^R} \in \mathds{R}^3$ is the interaction force between the telemanipulator and the environment and $\bm{\varOmega} \in \mathds{R}^{3x3}$ is a telefunctioning matrix defining the desired force relation between the telemanipulator and the foot master device.

The friction is assumed to be low. On the other hand, we expect Coriolis and Centrifugal forces to be negligible in the motion bandwidth of the leg. Hence, based on (\ref{eq_11}) the inverse dynamics can be approximated by: 
\begin{equation} \label{eq_13}
\bm{\tau_{u}^P \approx - b(q^P)\ddot{q}^P + g(q^P) + J_p^{T} \varOmega F_d^R}
\end{equation}
At the same time, each actuator's commanded torque (element $\tau_{u_i}$ of $\bm{\tau_u^P}$) is controlled through the current as  $\tau_{{u}_i} = k_{\tau} \cdotp i$, where $k_{\tau}$ is the torque constant  and $i$ is the current applied to the motor, tracked with a PI controller.


\section{Experimental Validation}

An experiment with the feet platforms was performed with two KUKA LWR IV+. The goals were to: 1. Evaluate the force transparency of the teleoperation (environment force vs foot interaction force) when the master device is used with implicit force control (no force sensor used in closed loop), 2. Assess the DS modulated impedance control in terms of effects on position-force tracking in the teleoperation. 3. Check the feasibility of a bipedal grasp in cartesian motion.

A volunteer from the research group (main author) performed the test. He got familiarized with the device for 20 minutes before the task, moving around their feet and observing the behaviour in the robot arms. Indeed,the calibration of the tele-functioning and the robot impedance matrix was based on what was perceived to be comfortable by the user. Namely, the reflected force in $z$ (axis of gravity) was scaled down five times for the master, while the components for $x$ and $y$ remained unchanged. Similarly, the platform's position ( in all axes ) was amplified five times from the master device to the robot's workspace.

\subsection{The Task}

The user was tasked to control the two robotic arms with coordinated and uncoordinated maneuvers. The task is illustrated and explained in detail in Fig. \ref{fig_snapshot}. 
\subsection{Results}

Fig. \ref{fig_human_response_task} shows the temporal evolution of the position of ipsilateral right robotic arm and foot, as well as the interaction force in the environment and in the foot interface (N.B. The left side information is not included since at the moment of the experiment the left platform didn't have a force sensor). 

Plot $III$ indicates that the tracking error of the position in the direction of grasping  was lower in phases a,b and f (i.e. free motion), with a Root Mean Squared Error ($RMSE_y$) of $ 0.087 \text{ m } $, than in phases c-e (i.e contact) with a $RMSE_y = 0.107 \text{ m}$. Probably when trying to squeeze the object (phase e) the moving attractors controlling the position of the robotic arms, are virtually pushed further inside the object but the real walls of the object prevent convergence in this direction. This translates in a bigger error in position tracking. In contrast, regarding the orthogonal directions (c.f. plot $II$ and $IV$), the tracking error was greater in free motion ($ RMSE_x = 0.132 \text{ m, } RMSE_z = 0.074 \text{ m}$) than in contact ($ RMSE_x = 0.050 \text{ m, } RMSE_z = 0.060 \text{ m}$), because when trying to hold the object in place, the convergence to the attractor in these directions is not constrained. 

It is clear how the impedance endowed to the robot arm smooths down its motion, acting as a low pass filter in the position tracking. This translates in less jerky movements (see plot $II$ phase b). Indeed, the task-aligned gain of would have to be tuned depending on the required task to find a trade-off between compliance and accuracy. On the other hand, regarding the directions orthogonal to the task, the damping seems to contribute to the stability of the grasp and to low startling of the subject (evidenced in low reactivity in $X$ and $Z$ for the human motion during phases c-e) specially during the abrupt disturbances (i.e. hammering).

Results show a low error ($RMSE_y = 0.033 \text{ N}$) in force reflection during contact (i.e. phases c-e). This means that the foot platforms are very transparent. Such result can be attributed to the high backdrivability (due to low gear ratio employed) and the smooth motion of the joints chosen for the mechanical construction.

To conclude, the task of grasping, lifting, working on an object, and overcoming abrupt perturbations with the feet was found feasible and successfully achieved.

\section{Summary and Outlook}

The contribution presented in this paper evaluates the use of the feet for direct control of robotic arms to be used along with the biological hands in a manipulation task. Both an experimental prototype and a control implementation are presented along with a preliminary demonstration. Results show a human being able to perform and maintain a bipedal grasp with high force transparency in the task-aligned direction and rejection of abrupt disturbances in the orthogonal directions to the task. These selectively convenient behaviours were possible thanks to the control strategy adopted (Impedance Control modulated through a Dynamical System).

Despite of using an open-loop implicit haptic control with approximative compensation of the dynamics, the force error during contact resulted to be was small. This initially validates good mechanics  for haptics and discourages the need for closed loop force control. Nevertheless, further characterization on the platforms regarding Z-width and friction identification should be done.

We are currently putting effort in evaluations with more participants and definition of protocols for training and determination of subject-specific calibration parameters. Additionally, a new version of the platform with the same kinematics but with fully motorized 5 DoF is under development.

For the next steps, we focus in defining and testing more complex and concrete manipulation tasks with simultaneous four arm interactions. On the other hand, it is likely that controlling four arms simultaneously may create an additional cognitive load for the human. To alleviate this, we are addressing investigations on autonomy for the robots to control the two robotic arms in coordination so as to synchronize motion and force. Specifically, moving up in the spectrum of shared autonomy, the next immediate step is to combine the direct control presented in this paper, with a dynamical system's approach for motion and force generation in contact tasks (c.f. \cite{Billard-RSS-19})

\section{Acknowledgement}
We thank the support of the Hasler Foundation and the  European Community Horizon 2020, in particular the robotics program ICT-23-2014 under grant agreement 644727-CogIMon.

\bibliographystyle{aaai}
\bibliography{ai_hri_2019_12}

\begin{thebibliography}{}

\bibitem[\protect\citeauthoryear{Abbink and van~der Helm}{2004}]{Abbink}
Abbink, D., and van~der Helm, F.
\newblock 2004.
\newblock {Force perception measurements at the foot}.
\newblock In {\em 2004 IEEE Int. Conf. Syst. Man Cybern. (IEEE Cat.
  No.04CH37583)}, volume~3,  2525--2529.
\newblock IEEE.

\bibitem[\protect\citeauthoryear{Abdi \bgroup et al\mbox.\egroup
  }{2016}]{Abdi2016}
Abdi, E.; Burdet, E.; Bouri, M.; Himidan, S.; and Bleuler, H.
\newblock 2016.
\newblock {In a demanding task, three-handed manipulation is preferred to
  two-handed manipulation}.
\newblock {\em Sci. Rep.}

\bibitem[\protect\citeauthoryear{Abdi \bgroup et al\mbox.\egroup
  }{2017}]{Abdi2017a}
Abdi, E.; Bouri, M.; Burdet, E.; Himidan, S.; and Bleuler, H.
\newblock 2017.
\newblock {Positioning the endoscope in laparoscopic surgery by foot:
  Influential factors on surgeons' performance in virtual trainer}.
\newblock In {\em Proc. Annu. Int. Conf. IEEE Eng. Med. Biol. Soc. EMBS}.

\bibitem[\protect\citeauthoryear{Amanhoud, Khoramshahi, and
  Billard}{2019}]{Billard-RSS-19}
Amanhoud, W.; Khoramshahi, M.; and Billard, A.
\newblock 2019.
\newblock A dynamical system approach to motion and force generation in contact
  tasks.
\newblock In {\em Proceedings of Robotics: Science and Systems}.

\bibitem[\protect\citeauthoryear{ATI}{2018}]{ATI_S2018}
ATI, I.~A.
\newblock 2018.
\newblock {ATI Industrial Automation: F/T Sensor Mini40}.

\bibitem[\protect\citeauthoryear{Bonilla and Asada}{2014}]{Bonilla2014}
Bonilla, B.~L., and Asada, H.~H.
\newblock 2014.
\newblock {A robot on the shoulder: Coordinated human-wearable robot control
  using Coloured Petri Nets and Partial Least Squares predictions}.
\newblock In {\em Proc. - IEEE Int. Conf. Robot. Autom.}

\bibitem[\protect\citeauthoryear{Bright and {Harry Asada}}{2017}]{Bright2017}
Bright, Z., and {Harry Asada}, H.
\newblock 2017.
\newblock {Supernumerary Robotic Limbs for Human Augmentation in Overhead
  Assembly Tasks}.
\newblock In {\em {Supernumerary Robotic Limbs for Human Augmentation in
  Overhead Assembly Tasks}}.

\bibitem[\protect\citeauthoryear{Dettwyler \bgroup et al\mbox.\egroup
  }{2004}]{Dettwyler2004}
Dettwyler, M.; Stacoff, A.; {Kramers-de Quervain}, I.~A.; and St{\"{u}}ssi, E.
\newblock 2004.
\newblock {Modelling of the ankle joint complex. Reflections with regards to
  ankle prostheses}.
\newblock {\em Foot Ankle Surg.} 10(3):109--119.

\bibitem[\protect\citeauthoryear{Edworthy and Stanton}{1995}]{EDWORTHY1995}
Edworthy, J., and Stanton, N.
\newblock 1995.
\newblock {A user-centred approach to the design and evaluation of auditory
  warning signals: 1. Methodology}.
\newblock {\em Ergonomics} 38(11):2262--2280.

\bibitem[\protect\citeauthoryear{Farjadian \bgroup et al\mbox.\egroup
  }{2014}]{Farjadian2014}
Farjadian, A.~B.; Suri, S.; Bugliari, A.; Doucot, P.; Lavins, N.; Mazzotta, A.;
  Valenzuela, J.~P.; Murphy, P.; Kong, Q.; Holden, M.~K.; and Mavroidis, C.
\newblock 2014.
\newblock {vi-RABT: Virtually Interfaced Robotic Ankle and Balance Trainer}.
\newblock In {\em 2014 IEEE Int. Conf. Robot. Autom.},  228--233.
\newblock IEEE.

\bibitem[\protect\citeauthoryear{Geitner \bgroup et al\mbox.\egroup
  }{2018}]{Geitner2018}
Geitner, C.; Birrell, S.; Krehl, C.; and Jennings, P.
\newblock 2018.
\newblock {Haptic Foot Pedal: Influence of Shoe Type, Age, and Gender on
  Subjective Pulse Perception}.
\newblock {\em Hum. Factors J. Hum. Factors Ergon. Soc.} 60(4):496--509.

\bibitem[\protect\citeauthoryear{Girone \bgroup et al\mbox.\egroup
  }{2001}]{Girone2001}
Girone, M.; Burdea, G.; Bouzit, M.; Popescu, V.; and Deutsch, J.~E.
\newblock 2001.
\newblock {Stewart platform-based system for ankle telerehabilitation}.
\newblock {\em Auton. Robots}.

\bibitem[\protect\citeauthoryear{Hoffman}{1991}]{HOFFMANN1991}
Hoffman, E.~R.
\newblock 1991.
\newblock {A comparison of hand and foot movement times}.
\newblock {\em Ergonomics} 34(4):397--406.

\bibitem[\protect\citeauthoryear{Ichinose, Gomikawa, and
  Suzuki}{2013}]{Ichinose2013}
Ichinose, A.; Gomikawa, Y.; and Suzuki, S.
\newblock 2013.
\newblock {Driving assistance through pedal reaction force control with
  consideration of JND}.
\newblock In {\em 2013 IEEE RO-MAN},  484--489.
\newblock IEEE.

\bibitem[\protect\citeauthoryear{Iwata, Yano, and Nakaizumi}{2001}]{Iwata2001}
Iwata, H.; Yano, H.; and Nakaizumi, F.
\newblock 2001.
\newblock {Gait Master: a versatile locomotion interface for uneven virtual
  terrain}.
\newblock In {\em Proc. IEEE Virtual Real. 2001},  131--137.
\newblock IEEE Comput. Soc.

\bibitem[\protect\citeauthoryear{Kronander and Billard}{2016}]{Kronander2016}
Kronander, K., and Billard, A.
\newblock 2016.
\newblock {Passive Interaction Control With Dynamical Systems}.
\newblock {\em IEEE Robot. Autom. Lett.} 1(1):106--113.

\bibitem[\protect\citeauthoryear{Llorens-Bonilla, Parietti, and
  Asada}{2012}]{Llorens-Bonilla2012}
Llorens-Bonilla, B.; Parietti, F.; and Asada, H.~H.
\newblock 2012.
\newblock {Demonstration-based control of supernumerary robotic limbs}.
\newblock In {\em IEEE Int. Conf. Intell. Robot. Syst.}

\bibitem[\protect\citeauthoryear{Otis \bgroup et al\mbox.\egroup
  }{2008}]{Otis2008}
Otis, M. J.-D.; Mokhtari, M.; {Du Tremblay}, C.; Laurendeau, D.; {De
  Rainville}, F.~M.; and Gosselin, C.~M.
\newblock 2008.
\newblock {Hybrid control with multi-contact interactions for 6DOF haptic foot
  platform on a cable-driven locomotion interface}.
\newblock In {\em Symp. Haptics Interfaces Virtual Environ. Teleoperator Syst.
  2008 - Proceedings, Haptics},  161--168.
\newblock IEEE.

\bibitem[\protect\citeauthoryear{Pakkanen and Raisamo}{2004}]{Pakkanen2004}
Pakkanen, T., and Raisamo, R.
\newblock 2004.
\newblock {Appropriateness of foot interaction for non-accurate spatial tasks}.
\newblock In {\em Ext. Abstr. 2004 Conf. Hum. factors Comput. Syst. - CHI '04},
   1123.
\newblock New York, New York, USA: ACM Press.

\bibitem[\protect\citeauthoryear{Paradiso \bgroup et al\mbox.\egroup
  }{2004}]{Paradiso2004}
Paradiso, J.~A.; Morris, S.~J.; Benbasat, A.~Y.; and Asmussen, E.
\newblock 2004.
\newblock {Interactive therapy with instrumented footwear}.
\newblock In {\em Ext. Abstr. 2004 Conf. Hum. factors Comput. Syst. - CHI '04},
   1341.
\newblock New York, New York, USA: ACM Press.

\bibitem[\protect\citeauthoryear{Pheasant}{1996}]{Pheasant1996}
Pheasant, S.
\newblock 1996.
\newblock {\em {Bodyspace : anthropometry, ergonomics, and the design of
  work}}.
\newblock Taylor {\&} Francis.

\bibitem[\protect\citeauthoryear{Plagenhoef, Evans, and
  Abdelnour}{1983}]{Plagenhoef1983}
Plagenhoef, S.; Evans, F.~G.; and Abdelnour, T.
\newblock 1983.
\newblock {Anatomical Data for Analyzing Human Motion}.
\newblock {\em Res. Q. Exerc. Sport} 54(2):169--178.

\bibitem[\protect\citeauthoryear{Rovers and van Essen}{2005}]{Rovers2005}
Rovers, A., and van Essen, H.
\newblock 2005.
\newblock {FootIO: Design and Evaluation of a Device to Enable Foot Interaction
  over a Computer Network}.
\newblock In {\em First Jt. Eurohaptics Conf. Symp. Haptic Interfaces Virtual
  Environ. Teleoperator Syst.},  521--522.
\newblock IEEE.

\bibitem[\protect\citeauthoryear{Saglia \bgroup et al\mbox.\egroup
  }{2013}]{Saglia2013}
Saglia, J.~A.; Tsagarakis, N.~G.; Dai, J.~S.; and Caldwell, D.~G.
\newblock 2013.
\newblock {Control Strategies for Patient-Assisted Training Using the Ankle
  Rehabilitation Robot (ARBOT)}.
\newblock {\em IEEE/ASME Trans. Mechatronics} 18(6):1799--1808.

\bibitem[\protect\citeauthoryear{Siegler, Chen, and
  Schneck}{1988}]{Siegler1988}
Siegler, S.; Chen, J.; and Schneck, C.~D.
\newblock 1988.
\newblock {The three-dimensional kinematics and flexibility characteristics of
  the human ankle and subtalar joints--Part I: Kinematics.}
\newblock {\em J. Biomech. Eng.} 110(4):364--73.

\bibitem[\protect\citeauthoryear{Southall}{1985}]{Southall1985}
Southall, D.
\newblock 1985.
\newblock {The discrimination of clutch-pedal resistances}.
\newblock {\em Ergonomics} 28(9):1311--1317.

\bibitem[\protect\citeauthoryear{Wang \bgroup et al\mbox.\egroup
  }{2013}]{Wang2013}
Wang, C.; Fang, Y.; Guo, S.; and Chen, Y.
\newblock 2013.
\newblock {Design and Kinematical Performance Analysis of a 3- R US/ R RR
  Redundantly Actuated Parallel Mechanism for Ankle Rehabilitation}.
\newblock {\em J. Mech. Robot.} 5(4):041003.

\bibitem[\protect\citeauthoryear{Yoon and Ryu}{2006}]{Yoon2006a}
Yoon, J., and Ryu, J.
\newblock 2006.
\newblock {A Novel Locomotion Interface with Two 6-DOF Parallel Manipulators
  That Allows Human Walking on Various Virtual Terrains}.
\newblock {\em Int. J. Rob. Res.} 25(7):689--708.

\end{thebibliography}

\end{document}